\documentclass[9.5pt,journal,final,finalsubmission,twocolumn]{IEEEtran}
\usepackage{graphicx}
\usepackage{amsmath,amssymb} 

\newcommand{\etal}{\mbox{\emph{et al.} }}
\usepackage{multirow,tabularx}
\usepackage{hyperref}

\begin{document}
\title{Crafting GBD-Net for Object Detection}
\author{  Xingyu~Zeng*,%~\IEEEmembership{Student Member,~IEEE, }
Wanli~Ouyang*,%~\IEEEmembership{Member,~IEEE,} 
Junjie~Yan,  Hongsheng~Li,%~\IEEEmembership{Member,~IEEE},  
Tong~Xiao,  Kun~Wang, Yu Liu, Yucong Zhou, Bin Yang, 
Zhe~Wang,%~\IEEEmembership{Student Member,~IEEE}, 
 Hui~Zhou,  
 Xiaogang~Wang,%~\IEEEmembership{Member,~IEEE},  
%\thanks{Manuscript received ...; revised ....}% <-this % stops a space
\thanks{ Xingyu~Zeng (equal contribution), Wanli Ouyang (equal contribution), Tong Xiao, Kun Wang, Hongsheng~Li, Zhe Wang, Hui Zhou and Xiaogang Wang are with the Department of Electronic Engineering at the Chinese University of Hong Kong, Hong Kong. Junjie Yan,Yu Liu, Yucong Zhou, Bin Yang are with the Sensetime Group Limited.}}
% note the % following the last \IEEEmembership and also the first \thanks -
% instance, "A\textbf{} \textbf{}B" will typeset as "A B" not "AB". If you want
% "AB" then you have to do: "A\textbf{}\textbf{}B"
\markboth{Manuscript}{Shell \MakeLowercase{\textit{et al.}}: Bare Demo of IEEEtran.cls for Journals}

% If you want to put a publisher's ID mark on the page
% (can leave text blank if you just want to see how the
% text height on the first page will be reduced by IEEE)
%\pubid{0000--0000/00\$00.00~\copyright~2002 IEEE}

%\specialpapernotice{(Invited Paper)}

\maketitle

\begin{abstract}
The visual cues from multiple support regions of different sizes and resolutions are complementary in classifying a candidate box in object detection. Effective integration of local and contextual visual cues from these regions has become a fundamental problem in object detection. %Most existing works simply concatenated features or scores obtained from support regions.
 In this paper, we propose a gated bi-directional CNN (GBD-Net) to pass messages among features from different support regions during both feature learning and feature extraction. Such message passing can be implemented through convolution between neighboring support regions in two directions and can be conducted in various layers. Therefore, local and contextual visual patterns can validate the existence of each other by learning their nonlinear relationships and their close interactions are modeled in a more complex way. It is also shown that message passing is not always helpful but dependent on individual samples. Gated functions are therefore needed to control message transmission, whose on-or-offs are controlled by extra visual evidence from the input sample. The effectiveness of GBD-Net is shown through experiments on three object detection datasets, ImageNet, Pascal VOC2007 and Microsoft COCO. This paper also shows the details of our approach in wining the ImageNet object detection challenge of 2016, with source code provided on \url{https://github.com/craftGBD/craftGBD}.
\end{abstract}

\begin{IEEEkeywords}
Convolutional neural network, CNN, deep learning, deep model, object detection.
\end{IEEEkeywords}

\section{Introduction}
Object detection is one of the fundamental vision problems. It provides basic information for semantic understanding of images and videos. Therefore, it has attracted a lot of attentions. Detection is regarded as a problem of classifying candidate boxes. Due to large variations in viewpoints, poses, occlusions, lighting conditions and background, object detection is challenging.
Recently, since the seminal work in \cite{girshick2014rich}, convolutional neural networks (CNNs) \cite{LeCun:CNNDoc, Krizhevsky:ImageNetCNN,sermanet2013overfeat,simonyan2014very,szegedy2015going} have been proved to be effective for object detection  because of its power in learning features.

\begin{figure}
\begin{center}
\includegraphics[width=1\linewidth]{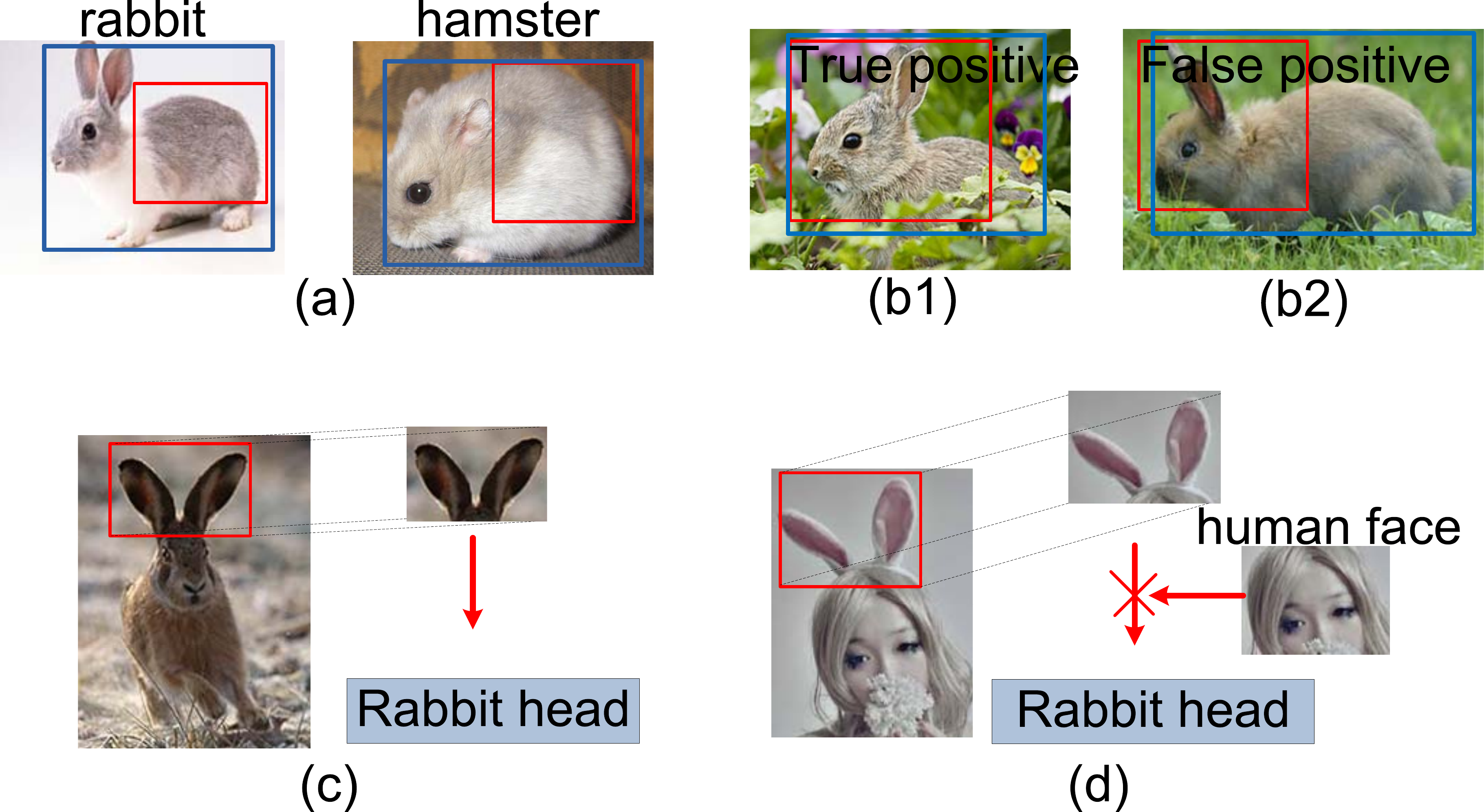}
\end{center}
\caption{The necessity of passing messages among features from supporting regions of different resolutions, and  controlling message passing according different image instances. Blue windows indicate the ground truth bounding boxes. Red windows are candidate boxes. It is hard to classify candidate boxes which cover parts of objects because of similar local visual cues in (a) and ignorance of the occlusion in (b). Local details of rabbit ears are useful for recognizing the rabbit head in (c). The contextual human head helps to find that the rabbit ear worn on human head should not be used to validate the existence of the rabbit head in (d). Best viewed in color.}
\label{fig:figure1}
\end{figure}

In object detection, a candidate box is counted as true-positive for an object category if the intersection-over-union (IoU) between the candidate box and the ground-truth box is greater than a threshold.
When a candidate box covers only  a part of the./ ground-truth regions, there are some potential problems.
\begin{itemize}
\item Visual cues in this candidate box may not be sufficient to distinguish object categories. Take the candidate boxes in Fig. \ref{fig:figure1}(a) for example, they cover parts of animal bodies and have similar visual cues, but with different ground-truth class labels. It is hard to distinguish their class labels without information from larger surrounding regions of the candidate boxes. \\
\item Classification on the candidate boxes depends on how much an object is occluded, which has to be inferred from larger surrounding regions. Because of occlusion, the candidate box covering a rabbit head in Fig. \ref{fig:figure1}(b1) should be considered as a true positive of rabbit, because of large IoU with the ground truth. Without occlusion, however, the candidate box covering a rabbit head in Fig. \ref{fig:figure1}(b2) should \textbf{not} be considered as a true positive because of small IoU with the ground truth. \\
\end{itemize}
To handle these problems, contextual regions surrounding candidate boxes are naturally helpful. Besides, surrounding regions also provide contextual information about background and other nearby objects to help detection. Therefore, in our deep model design and some existing works \cite{farabet2013learning}, information from surrounding regions are used to improve classification of a candidate box.

On the other hand, when CNN takes a large region as input, it sacrifices the ability in describing local details, which are sometimes critical in discriminating object classes, since CNN encodes input to a fixed-length feature vector. For example, the sizes and shapes of ears are critical details in discriminating rabbits from hamsters. But they may not be identified when they are in a very small part of the CNN input. It is desirable to have a network structure that takes both surrounding regions and local part regions into consideration.
Besides, it is well-known that features from different resolutions are complementary  \cite{farabet2013learning}.

One of our motivations is that features from different resolutions and support regions validate the existence of one another. For example, the existence of rabbit ears in a local region helps to strengthen the existence of a rabbit head, while the existence of the upper body of a rabbit in a larger contextual region also helps to validate the existence of a rabbit head. Therefore, we propose that features with different resolutions and support regions should pass messages to each other in multiple layers in order to validate their existences jointly during both feature learning and feature extraction. This is different from the naive way of learning a separate CNN for each support region and concatenating feature vectors or scores from different support regions for classification. 

Our further motivation is that care should be taken when passing messages among contextual and local regions. The messages are not always useful.
Taking Fig. \ref{fig:figure1}(c) as an example, the local details of the rabbit ear is helpful in recognizing the rabbit head, and therefore, its existence has a large weight in determining the existence of the rabbit head. However, when this rabbit ear is artificial and worn on a girl's head in Fig. \ref{fig:figure1}(d), it should not be used as the evidence to support the existence of a rabbit head. 
Extra information is needed to determine whether the message of a contextual visual pattern, e.g. rabbit ear, should be transmitted to support the existence of a target visual pattern, e.g. rabbit head. In Fig.  \ref{fig:figure1}(d), for example, the extra human-face visual cues indicates that the message of the rabbit ear should not be transmitted to strengthen the evidence of seeing the rabbit head. Taking this observation into account, we design a network that uses extra information from the input image region to adaptively control message transmission.

In this paper, we propose a gated bi-directional CNN (GBD-Net) architecture that adaptively models interactions of contextual and local visual cues during feature learning and feature extraction. Our contributions are in three-fold.
\begin{itemize}
\item A bi-directional network structure is proposed to pass messages among features from multiple support regions of different resolutions. With this design, local patterns pass detailed visual messages to larger patterns and large patterns passes contextual visual messages in both directions. Therefore, local and contextual features cooperate with each other on improving detection accuracy. We implement message passing by convolution.
\item We propose to control message passing with gate functions. With such gate functions, message from a found pattern is transmitted only when it is useful for some samples, but is blocked for others. 
\item A new deep learning pipeline for object detection. It effectively integrates region proposal, feature representation learning, context modeling, and model averaging into the detection system. Detailed component-wise analysis is provided through extensive experimental evaluation. This paper also investigates the influence of CNN structures for the large-scale object detection task under the same setting. The details of our submission to  the ImageNet Object Detection Challenge is provided in this paper, with source code provided online.
\end{itemize}
The proposed GBD-Net is implemented under the fast RCNN detection frameworks \cite{girshick2015fast}. The effectiveness is validated through the experiments on three datasets, ImageNet \cite{ILSVRCarxiv14}, PASCAL VOC2007 \cite{Everingham:PacalVOC} and Microsoft COCO \cite{lin2014microsoft}.

\section{Related work}
Impressive improvements have been achieved on object detection in recent years. They  mainly come from better region proposals, detection pipeline, feature learning algorithms and CNN structures, iterative bounding box regression and making better use of local and contextual visual cues.

\textbf{Region proposal.} Selective search \cite{Smeulders:SelectiveSearch} obtained region proposals by hierarchically grouping segmentation results. Edgeboxes \cite{ZitnickDollarECCV14edgeBoxes} evaluated the number of contours enclosed by a bounding box to indicate the likelihood of an object. Faster RCNN \cite{ren2015faster} and CRAFT \cite{yang2016craft} obtained region proposals with the help of convolutional neural networks. Pont-Tuest and Van Gool \cite{pont2015boosting} studied the statistical difference between the Pascal-VOC dataset \cite{Everingham:PacalVOC} to Microsoft CoCo dataset \cite{lin2014microsoft} to obtain better object proposals. Region proposal can be considered as a cascade of classifiers. At the first stage, millions of windows that are highly confident of being background are removed by region proposal methods and the remaining hundreds or southands are then used for classification. In this paper, we adopt an improved version of the CRAFT in providing the region proposals.

\textbf{Iterative regression.} Since the candidate regions are not very accurate in locating objects, multi-region CNN \cite{gidaris2015object}, LocNet \cite{gidaris2016locnet} and AttractioNet \cite{gidaris2016attend} are proposed for more accurate localization of the objects. These approaches conduct bounding box regression iteratively so that the candidate regions gradually move  towards the ground truth object.

\textbf{Object detection pipeline.} The state-of-the-art deep learning based object detection pipeline RCNN \cite{girshick2014rich} extracted CNN features from the warped image regions and applied a linear SVM as the classifier. By pre-training on the ImageNet classification dataset, it achieved great improvement in detection accuracy compared with previous sliding-window approaches that used handcrafted features on PASCAL-VOC and the large-scale ImageNet object detection dataset. In order to obtain a higher speed, Fast RCNN \cite{girshick2015fast} shared the computational cost among candidate boxes in the same image and proposed a novel roi-pooling operation to extract feature vectors for each region proposal. Faster RCNN \cite{ren2015faster} combined the region proposal step with the region classification step by sharing the same convolution layers for both tasks. Region proposal is not necessary. Some recent approaches, e.g. Deep MultiBox \cite{szegedy2014scalable},  YOLO \cite{redmon2015you} and SSD \cite{liu2016ssd}, directly estimate the object classes from predefined sliding windows. 

\textbf{Learning and design of CNN.} A large number of works \cite{Krizhevsky:ImageNetCNN,sermanet2013overfeat,simonyan2014very,szegedy2015going,zagoruyko2016multipath,he2016deep} aimed at designing network structures and they are shown to be effective in the detection task. The works in \cite{Krizhevsky:ImageNetCNN,sermanet2013overfeat,simonyan2014very,szegedy2015going,he2016deep} proposed deeper networks. People \cite{ioffe2015batch,simonyan2014very,he2015delving} also investigated how to effectively train deep networks. Simonyan \etal \cite{simonyan2014very} learn deeper networks based on the parameters in shallow networks. Ioffe \etal \cite{ioffe2015batch} normalized each layer inputs for each training mini-batch in order to avoid internal covariate shift. He \etal \cite{he2015delving} investigated parameter initialization approaches and proposed parameterized RELU. Li \etal \cite{li2016multi} proposed multi-bias non-linear activation (MBA) layer to explore the information hidden in the magnitudes of responses.

Our contributions focus on a novel bi-directional network structure to effectively make use of multi-scale and multi-context regions. Our design is complementary to above region proposals, pipelines, CNN layer designs, and training approaches.
There are many works on using visual cues from object parts \cite{ouyang2015deepid,gidaris2015object,girshick2015deformable} and contextual information \cite{ouyang2015deepid,gidaris2015object}. Gidaris \etal \cite{gidaris2015object} adopted a multi-region CNN model and manually selected multiple image regions. Girshick \etal \cite{girshick2015deformable} and Ouyang \etal \cite{ouyang2015deepid} learned the deformable parts from CNNs. In order to use the contextual information, multiple image regions surrounding the candidate box were cropped in \cite{gidaris2015object} and whole-image classification scores were used in \cite{ouyang2015deepid}. These works simply concatenated features or scores from object parts or context while we pass message among features representing local and contextual visual patterns so that they validate the existence of each other by non-linear relationship learning. Experimental results show that GBD-Net is complementary to the approaches in \cite{gidaris2015object, ouyang2015deepid}. As a step further, we propose to use gate functions for controlling message passing, which was not investigated in existing works.

\textbf{Passing messages and gate functions.} Message passing at the feature level was studied in Recurrent neural network (RNN) and gate functions are used to control message passing in long short-term memory (LSTM) networks. However, both techniques have not been used to investigate feature extraction from multi-resolution and multi-context regions yet, which is fundamental in object detection. Our message passing mechanism and gate functions are specifically designed for this problem setting. GBD-Net is also different from RCNN and LSTM in the sense that  it does not share parameters across resolutions/contexts. 

\section{Gated Bi-directional CNN}
\label{Sec:GBD}
We briefly introduce the fast RCNN pipeline in Section \ref{sec:FastRCNN} and then provide an overview of our approach in Section \ref{sec:overview}. 
Our use of roi-pooling is discussed in Section\ref{sec:input}. Section\ref{sec:net} focuses on the proposed bi-directional network structure and its gate function. Section \ref{Sec:GBD_ext} introduces the modified GBD structure. Section\ref{sec:training} explains the details of the training scheme.

\subsection{Fast RCNN pipeline}
\label{sec:FastRCNN}
We adopt the Fast RCNN\cite{girshick2015fast} as the object detection pipeline with four steps. 
\begin{itemize}
  \item Step 1) Candidate box generation. Thousands or hundreds of candidate boxes are selected from a large pool of boxes.
  \item Step 2) Feature map generation. Given an input as the input of CNN, feature maps are generated. 
  \item Step 3) Roi-pooling.  Each candidate box is considered as a region-of-interest (roi) and a pooling function is operated on the CNN feature maps generated in the step 2. After roi-pooling, candidate boxes of different sizes are pooled to have the same feature vector size.
  \item  Step 4) Classification. CNN features after roi-pooling go through several convolutions, pooling and fully connected layers to predict the class label and location refinement of candidate boxes.
\end{itemize}

\subsection{Framework overview}
\label{sec:overview}

\begin{figure*}[h]
\begin{center}
   \includegraphics[width=0.88\linewidth]{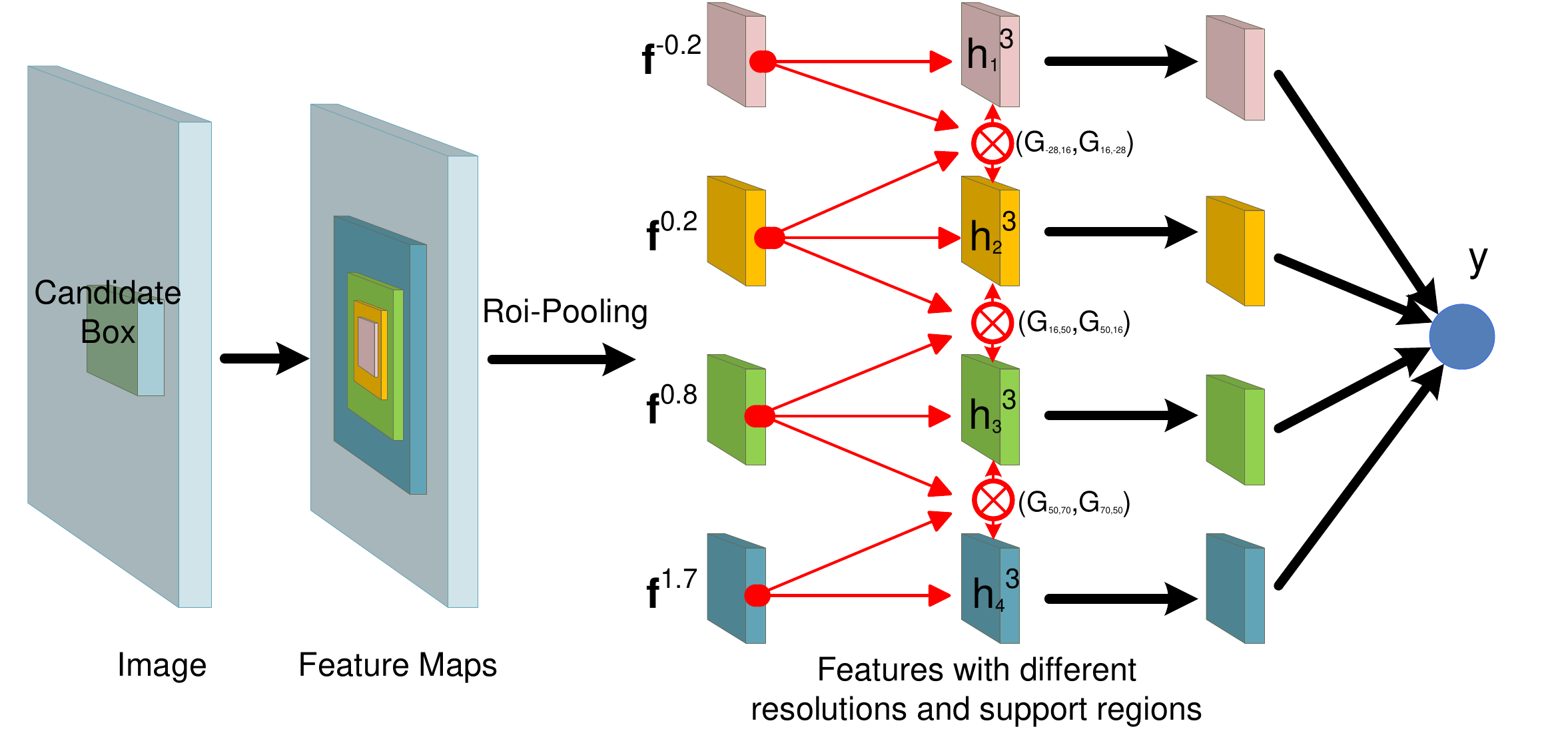}
\end{center}
   \caption{Overview of our framework. The network takes an image as input and produces feature maps. The roi-pooling is done on  feature maps to obtain features with different resolutions and support regions, denoted by $\mathbf{f}^{-0.2}, \mathbf{f}^{0.2}, \mathbf{f}^{0.8}$ and $\mathbf{f}^{1.7}$. Red arrows denote our gated bi-directional structure for passing messages among features. Gate functions $G$ are defined for controlling the message passing rate. Then all features $\mathbf{h}^3_i$ for $i = 1, 2,3, 4$ go through multiple CNN layers with shared parameters to obtain the final features that are used to predict the class $y$. Parameters on black arrows are shared across branches, while parameters on red arrows are not shared. Best viewed in color.}
\label{fig:overview}
\end{figure*}

An overview of the GBD-Net is shown in Fig. \ref{fig:overview}. Based on the fast RCNN pipeline, our proposed model takes an image as input, uses roi-pooling operations to obtain features with different resolutions and different support regions for each candidate box, and then the gated bi-direction layer is used for passing messages among features, and final classification is made. We use the BN-net \cite{ioffe2015batch} as the baseline network structure, i.e. if only one support region and one branch is considered,  Fig. \ref{fig:overview} becomes a BN-net. Currently, messages are passed between features in one layer. It can be extended by adding more layers between the roi-pooled feature $\mathbf{f}$ and refined feature $\mathbf{h}$ for
passing messages in these layers.

We use the same candidate box generation and feature map generation steps as the fast RCNN introduced in Section \ref{sec:FastRCNN}.
In order to take advantage of complementary visual cues in the surrounding/inner regions, the major modifications of fast RCNN are as follows.
\begin{itemize}
\item In the roi-pooling step, regions with the same center location but different sizes are pooled from the same feature maps for a single candidate box. The regions with different sizes before roi-pooling  have the same size after roi-pooling. In this way, the pooled features corresponds to different support regions and have different resolutions. 
\item Features with different resolutions optionally go through several CNN  layers to extract their high-level features.
\item The bi-directional structure is designed to pass messages among the roi-pooled features with different resolutions and support regions. In this way, features corresponding to different resolutions and support regions verify each other by passing messages to each other.
\item Gate functions are used to control message transmission. 
\item After message passing, the features for different resolutions and support regions are then passed through several CNN layers for classification.
\end{itemize}

\begin{figure*}[t]
\begin{center}
   \includegraphics[width=1\linewidth]{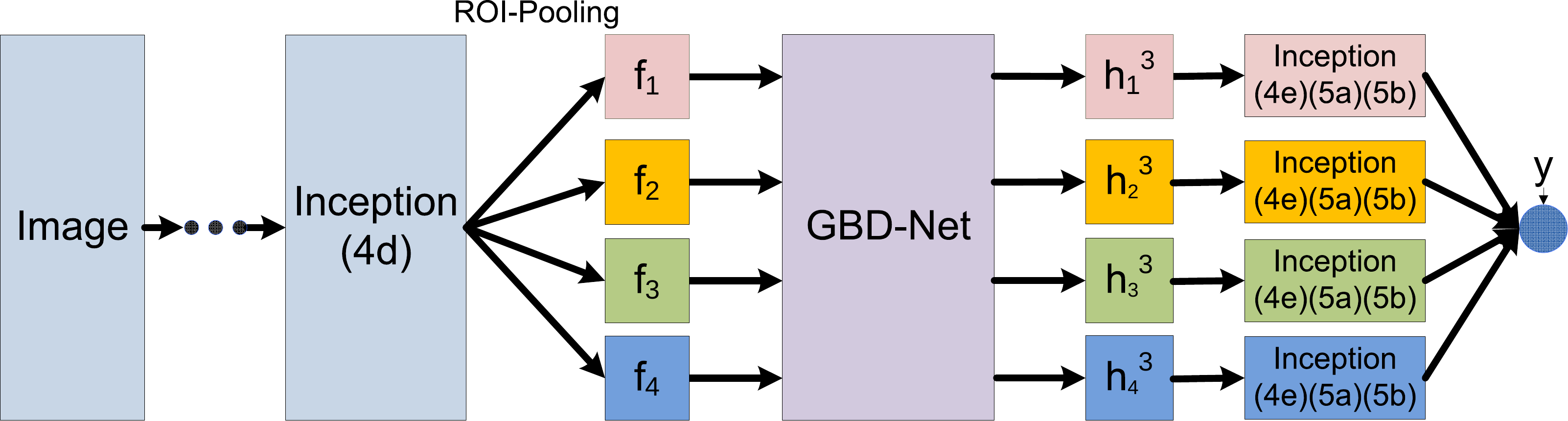}
\end{center}
   \caption{Exemplar implementation of our model. The gated bi-directional network, dedicated as GBD-Net, is placed between Inception (4d) and Inception (4e). Inception (4e),(5a) and (5b) are shared among all branches.}
\label{fig:examplar}
\end{figure*}

An exemplar implementation of our model is shown in Fig. \ref{fig:examplar}. 
There are 9 inception modules in the BN-net \cite{ioffe2015batch}. Roi-pooling of multiple resolutions and support regions is conducted after the 6th inception module, which is inception (4d). Then the gated bi-directional network is used for passing messages among features and outputs $\mathbf{h}_1^3$, $\mathbf{h}_2^3$, $\mathbf{h}_3^3$, and $\mathbf{h}_4^3$. After message passing, $\mathbf{h}_1^3$, $\mathbf{h}_2^3$, $\mathbf{h}_3^3$ and $\mathbf{h}_4^3$ go through the 7th, 8th, 9th inception modules and the average pooling layers separately and then used for classification. There is option to place roi-pooling and GBD-Net after different layers of the BN-net. In Fig. \ref{fig:examplar}, they are placed after inception (4e). In the experiment, we also tried to place them  after the input image.

\subsection{Roi-pooling of features with different resolutions and support regions}
\label{sec:input}

We use the roi-pooling layer designed in \cite{girshick2015fast} to obtain features with different resolutions and support regions.
Given a candidate box $\mathbf{b}^{o} = \left[ x^{o},y^{o},w^{o},h^{o}\right]$ with center location $(x^{o},y^{o})$, width $w^{o}$  and height $h^{o}$, its padded bounding box is denoted by $\mathbf{b}^p$. $\mathbf{b}^p$ is obtained by enlarging the original box $\mathbf{b}^o$ along both $x$ and $y$ directions with scale $p$ as follows:
\begin{align}
\mathbf{b}^p = \left[ x^{o},y^{o}, (1+p)w^{o}, (1+p) h^{o}  \right].
\end{align}

In RCNN \cite{girshick2014rich}, $p$ is 0.2 by default and the input to CNN is obtained by warping all the pixels in the enlarged bounding box $\mathbf{b}^p$ to a fixed size $w \times h$, where $w=h=224$ for the BN-net \cite{ioffe2015batch}. 
In fast RCNN \cite{girshick2015fast}, warping is done on feature maps instead of pixels. For a box $\mathbf{b}^{o}$, its corresponding feature box $\mathbf{b}^{f}$ on the feature maps is calculated and roi-pooling uses max pooling to convert the features in $\mathbf{b}^{f}$  to feature maps with a fixed size. 

\begin{figure*}
\begin{center}
   \includegraphics[width=1\linewidth]{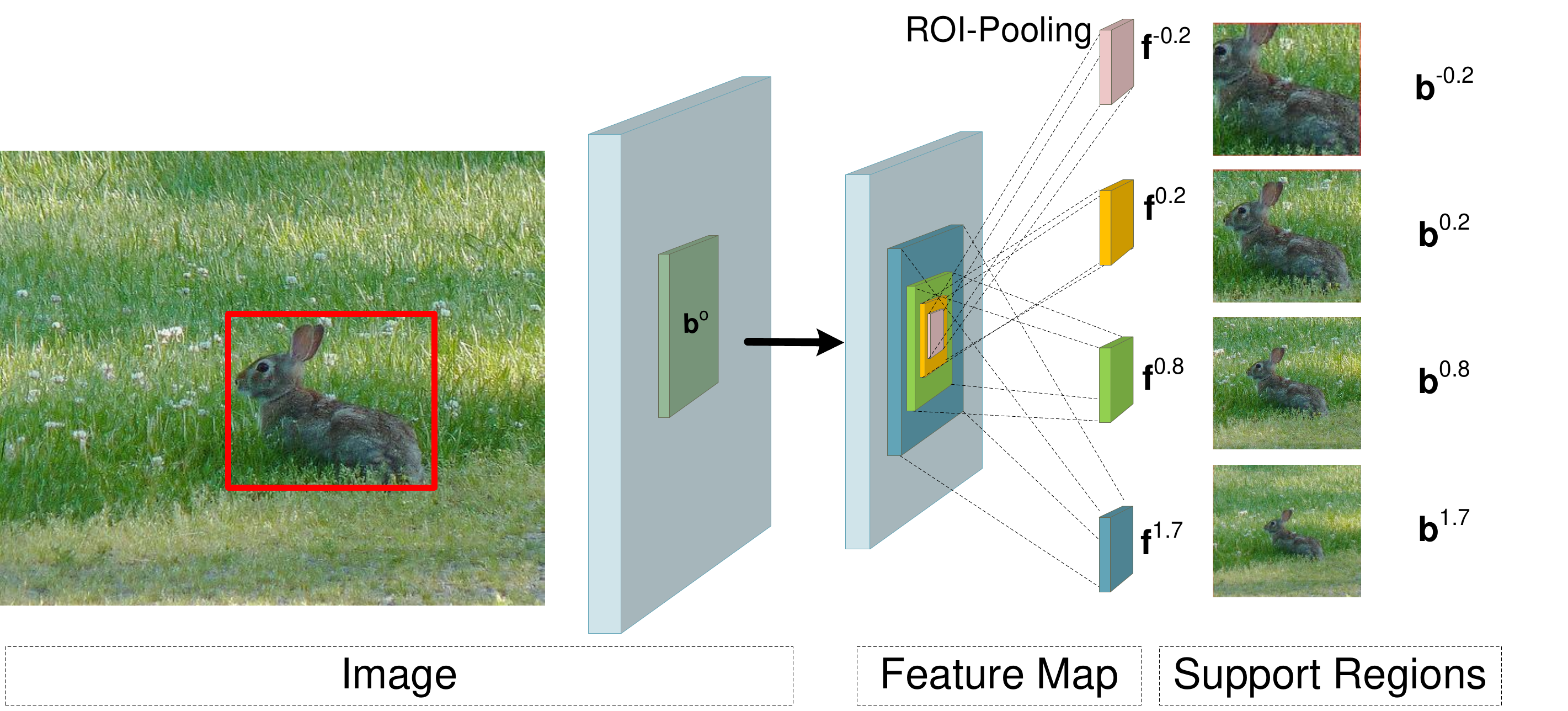}
\end{center}
   \caption{Illustration of using roi-pooling to obtain CNN features with different resolutions and support regions. The red rectangle in the left image is a candidate box. The right four image patches show the supporting regions for $\{\mathbf{b}^p\}$. Best viewed in color.}
\label{fig:input}
\end{figure*}

In our implementation, a set of padded bounding boxes $\{\mathbf{b}^p\}$ with different $p = -0.2, 0.2, 0.8, 1.7$  are generated for each candidate box $\mathbf{b}^{o}$. By roi-pooling on the CNN features, these boxes are warped into the same size, which is $14\times 14 \times 608$ for BN-net. The CNN features of these padded boxes have  different resolutions and support regions. In the roi-pooling step, regions corresponding to $\mathbf{b}^{-0.2}, \mathbf{b}^{0.2}, \mathbf{b}^{0.8}$ and $\mathbf{b}^{1.7}$ are warped into features $\mathbf{f}^{-0.2}, \mathbf{f}^{0.2}, \mathbf{f}^{0.8}$ and $\mathbf{f}^{1.7}$ respectively. Figure \ref{fig:input} illustrates this procedure.

Since features $\mathbf{f}^{-0.2}, \mathbf{f}^{0.2}, \mathbf{f}^{0.8}$ and $\mathbf{f}^{1.7}$ after roi-pooling are of the same size, the context scale value $p$ determines both the amount of padded context and also the resolution of the features. A larger $p$ value means a lower resolution for the original box but more contextual information around the original box, while a small $p$ means a higher resolution for the original box but less context.

\subsection{Gated Bi-directional network structure (GBD-v1)}
\label{sec:net}

\subsubsection{Bi-direction structure}
\begin{figure*}[ht]
\begin{center}
   \includegraphics[width=0.8\linewidth]{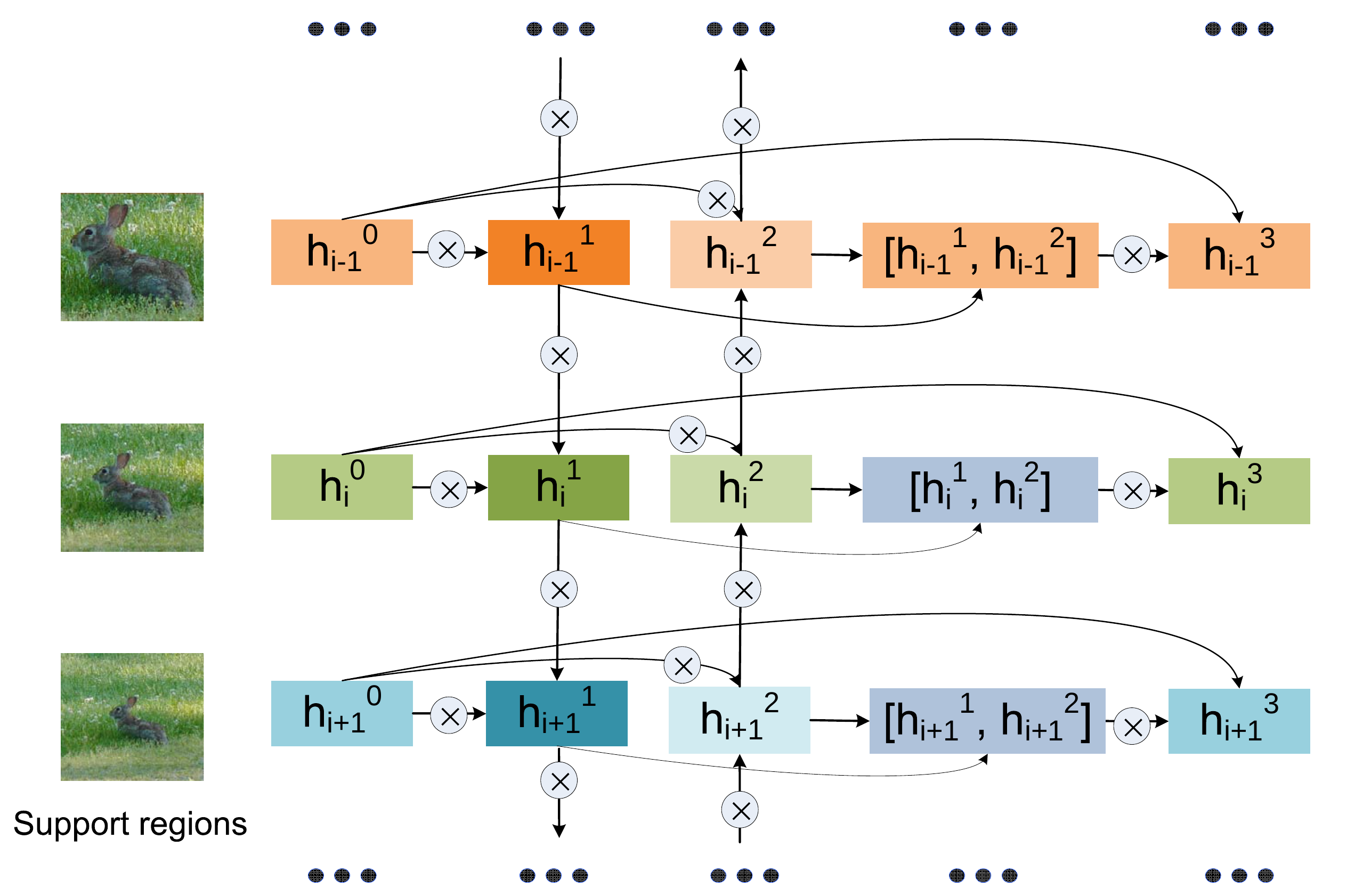}
\end{center}
   \caption{Details of our bi-directional structure. $\otimes$ denotes convolution. The input of this structure is the features $\{\mathbf{h}^0_{i}\}$ of multiple resolutions and contextual regions. Then bi-directional connections among these features are used for passing messages across resolutions/contexts. The output $\mathbf{h}^3_{i}$ are updated features for different resolutions/contexts after message passing.}
\label{fig:net}
\end{figure*}

Figure \ref{fig:net} shows the architecture of our proposed bi-directional network. 
It takes features $\mathbf{f}^{-0.2}, \mathbf{f}^{0.2}, \mathbf{f}^{0.8}$ and $\mathbf{f}^{1.7}$ as input and outputs features $\mathbf{h}_1^3$, $\mathbf{h}_2^3$, $\mathbf{h}_3^3$ and $\mathbf{h}_4^3$ for a single candidate box. In order to have features $\{\mathbf{h}_i^3\}$ with different resolutions and support regions cooperate with each other, this new structure builds two directional connections among them. One directional connection starts from features with the smallest region size and ends at features with the largest region size. The other is the opposite.

For a single candidate box $\mathbf{b}^{o}$,  $\mathbf{h}^0_{i}=\mathbf{f}^{p_i}$ representing features with context pad value $p_i$. 
The forward propagation for the proposed bi-directional structure can be summarized as follows:
\begin{align}
\mathbf{h}_{i}^{1} &= \sigma (\mathbf{h}^0_{i} \otimes \mathbf{w}_{i}^{1}+\mathbf{b}^{0,1}_{i}) + \sigma(\mathbf{h}_{i-1}^{1} \otimes \mathbf{w}_{i-1,i}^{1}+\mathbf{b}^1_{i}), \label{eqa:hi1} \\ 
& \tag{  high res. to low pass } \nonumber \\ 
\mathbf{h}_{i}^{2} &= \sigma(\mathbf{h}^0_{i} \otimes \mathbf{w}_{i}^{2}+\mathbf{b}^{0,2}_{i}) + \sigma(\mathbf{h}_{i+1}^{2} \otimes \mathbf{w}_{i,i+1}^{2}+\mathbf{b}^{2}_{i}), \label{eqa:hi2} \\
& \tag{low res. to high  pass} \nonumber \\
\mathbf{h}_{i}^{3} &=\sigma( cat(\mathbf{h}_{i}^{1},\mathbf{h}_{i}^{2}) \otimes \mathbf{w}_{i}^{3}+\mathbf{b}^{3}_{i}),  \label{eqa:hi3} \\
& \tag{  message integration}  \nonumber
\end{align}

\begin{itemize}
\item There are four different resolutions/contexts, $i = 1, 2, 3, 4$.
\item $\mathbf{h}^{1}_i$ represents the updated features after receiving message from $\mathbf{h}_{i-1}^{1}$ with a higher resolution and a smaller support region. It is assumed that $\mathbf{h}^{1}_0 = 0$, since $\mathbf{h}^{1}_1$ has the smallest support region and receives no message. 
\item $\mathbf{h}^{2}_i$ represents the updated features after receiving message from $\mathbf{h}_{i+1}^{2}$ with a lower resolution and a larger support region. It is assumed that $\mathbf{h}^{2}_5 = 0$, since $\mathbf{h}^{2}_4$ has the largest support region and receives no message.
\item $cat()$ concatenates CNN features maps along the channel direction.
\item The features $\mathbf{h}^{1}_i$ and $\mathbf{h}^{2}_i$ after message passing are integrated into $\mathbf{h}_{i}^{3}$  using the convolutional filters $\mathbf{w}_{i}^{3}$.

\item $\otimes$ represents the convolution operation. The biases and filters of convolutional layers are respectively denoted by $\mathbf{b}^{*}_{*}$ and $\mathbf{w}^{*}_{*}$. 
\item Element-wise RELU is used as the non-linear function $\sigma(\cdot)$.
\end{itemize}

From the equations above,  the features in $\mathbf{h}_{i}^{1}$ receive the messages from the high-resolution/small-context features and the features $\mathbf{h}_{i}^{2}$ receive messages from the low-resolution/large-context features. Then $\mathbf{h}_{i}^{3}$ collects messages from both directions to have a better representation of the $i$th resolution/context. For example, the visual pattern of a rabbit ear is obtained from features with a higher resolution and a smaller support region, and its existence (high responses in these features) can be used for validating the existence of a rabbit head, which corresponds to features with a lower resolution and a larger support region. This corresponds to message passing from high resolution to low resolution in (\ref{eqa:hi1}). 
Similarly, the existence of the rabbit head at the low resolution also helps to validate the existence of the rabbit ear at the high resolution by using (\ref{eqa:hi2}).  $\mathbf{w}_{i-1,i}^{1}$ and  $\mathbf{w}_{i,i+1}^{1}$ are learned to control how strong the existence of a feature with one resolution/context influences the existence of a feature with another resolution/context. 
Even after bi-directional message passing, $\{\mathbf{h}_i^3\}$ are complementary and will be jointly used for classification in later layers.

Our bi-directional structure is different from the bi-direction recurrent neural network (RNN). RNN aims to capture dynamic temporal/spatial behavior with a directed cycle. It is assumed that parameters are shared among directed connections. Since our inputs differ in both resolutions and contextual regions, convolutions layers connecting them should learn different relationships at different resolution/context levels. Therefore, the convolutional parameters for message passing are not shared in our bi-directional structure.

\subsubsection{Gate functions for message passing}
Instead of passing messages in the same way for all the candidate boxes, gate functions are introduced to adapt message passing for individual candidate boxes. Gate functions are also implemented as convolution. The design of gate filters considers the following aspects.

\begin{itemize}

\item $\mathbf{h}_i^k$ has multiple feature channels. A different gate filter is learned for each channel. 
\item The message passing rates should be controlled by the responses to particular visual patterns which are captured by gate filters.
\item The message passing rates can be determined by visual cues from nearby regions, e.g. in Fig. \ref{fig:figure1}, a girl's face indicates that the rabbit ear is artificial and should not pass message to the rabbit head. Therefore, the size of gate filters should not be $1 \times 1$ and $3 \times 3$ is used in our implementation. 
\end{itemize}

 \begin{figure*}
\begin{center}
   \includegraphics[width=0.8\linewidth]{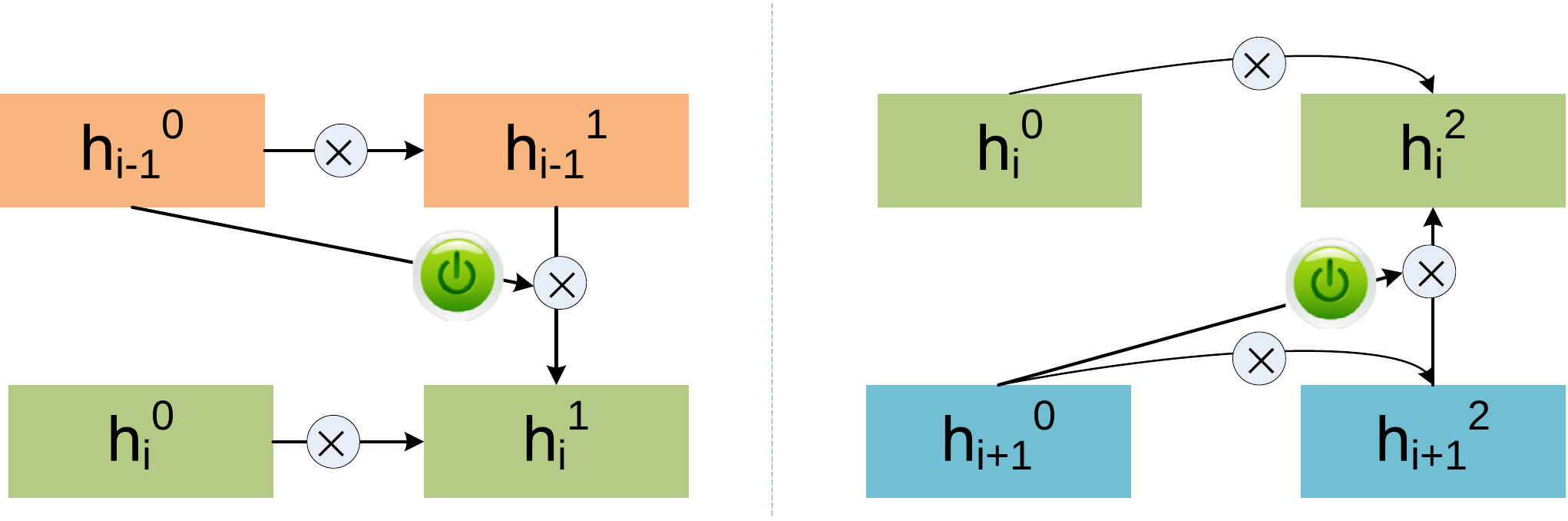}
\end{center}
   \caption{Illustration of the bi-directional structure with gate functions. The $\otimes$ represents the convolution and the switch button represents the gate function.}
\label{fig:gate}
\end{figure*}
We design gate functions as convolution layers with the sigmoid non-linearity to make the message passing rate in the range of (0,1). 
With gate functions, message passing in (\ref{eqa:hi1}) and (\ref{eqa:hi2}) for the bi-directional structure is changed:

\begin{align}
&\mathbf{h}_{i}^{1} = \sigma (\mathbf{h}^0_{i} \otimes \mathbf{w}_{i}^{1}+\mathbf{b}^{0,1}_{i}) +G_{i}^1  \bullet \sigma(\mathbf{h}_{i-1}^{1} \otimes \mathbf{w}_{i-1,i}^{1}+\mathbf{b}^1_{i}),  \label{eqa:hi5} \\ 
&\mathbf{h}_{i}^{2} = \sigma(\mathbf{h}^0_{i} \otimes \mathbf{w}_{i}^{2}+\mathbf{b}^{0,2}_{i}) +G_{i}^2 \bullet \sigma(\mathbf{h}_{i+1}^{2} \otimes \mathbf{w}_{i,i+1}^{2}+\mathbf{b}^{2}_{i}),  \label{eqa:hi6} \\
&G_{i}^1 = sigm(\mathbf{h}^0_{i-1} \otimes \mathbf{w}_{i-1,i}^{g}+\mathbf{b}_{i-1,i}^{g})  \label{eqa:gi1}\\
&G_{i}^2 =sigm(\mathbf{h}^0_{i+1}\otimes\mathbf{w}_{i+1,i}^{g}+\mathbf{b}_{i+1,i}^{g}) \label{eqa:gi2} 
\end{align}
\!\!\! where $sigm(\mathbf{x})=1/[1+\exp(-\mathbf{x})]$ is the element-wise sigmoid function and $\bullet$ denotes element-wise product. $G$ is the gate function to control message passing. It contains learnable convolutional parameters $\mathbf{w}^{g}_{*}, \mathbf{b}$ and uses features from the co-located regions to determine the rates of message  passing.  When $G(\mathbf{x},\mathbf{w}, \mathbf{b})$ is 0, the message is not passed.
The formulation for obtaining $\mathbf{h}_{i}^{3}$ is unchanged.  Fig. \ref{fig:gate} illustrates the bi-directional structure with gate functions. 

\subsection{The modified GBD structure (GBD-v2)}
\label{Sec:GBD_ext}
\begin{figure*}[]
\begin{center}
   \includegraphics[width=0.8\linewidth]{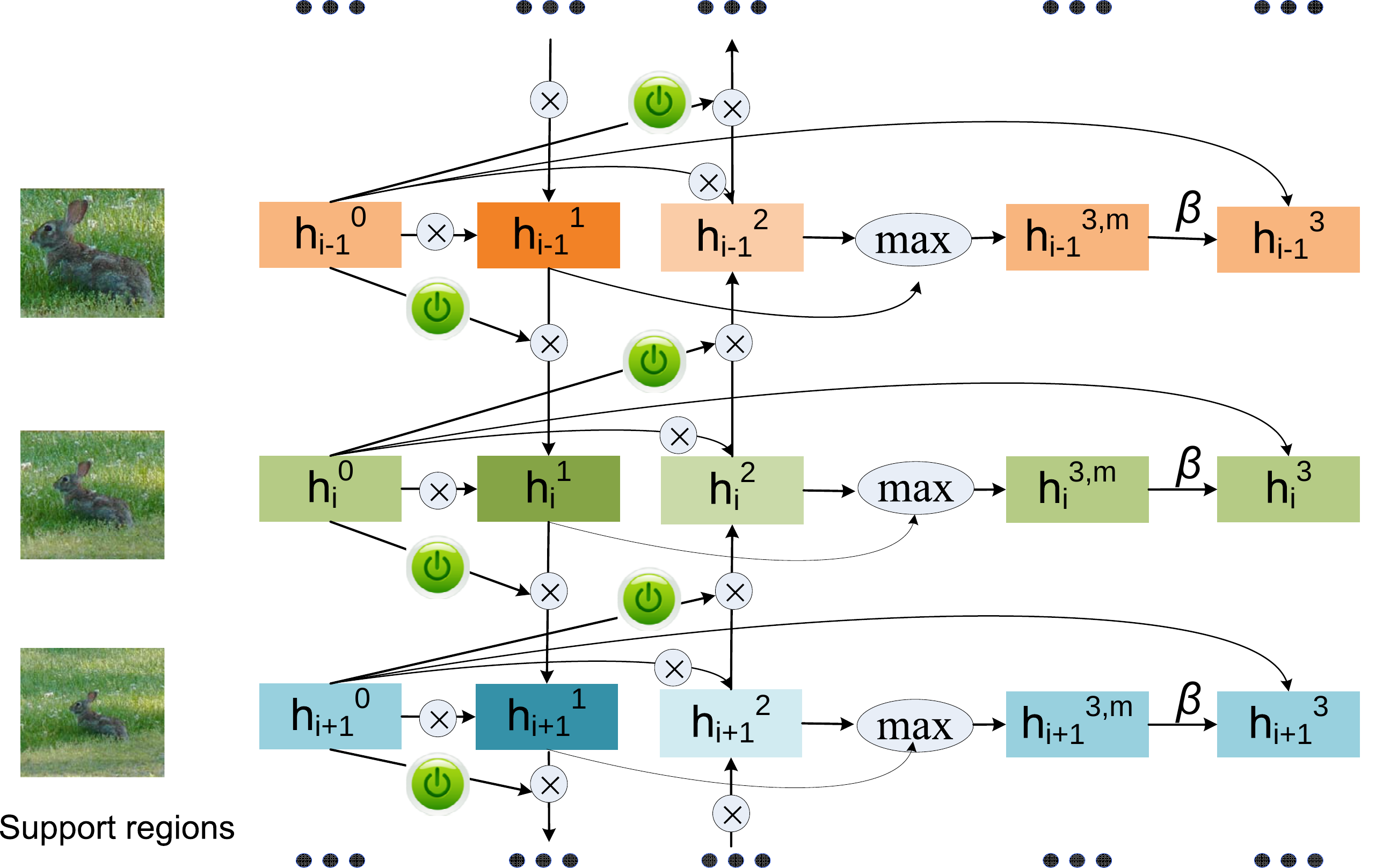}
\end{center}
   \caption{ Details of our modified bi-directional structure. Compared with the stucture in Fig. \ref{fig:net}, an identity mapping layer is added from $\mathbf{h}_*^0$ to $\mathbf{h}_*^3$.  The convolution from $[\mathbf{h}_*^1, \mathbf{h}_*^2]$ to  $\mathbf{h}_*^3$ in Fig. \ref{fig:net} is changed into max-pooling.}
\label{fig:net_ext}
\end{figure*}

For the models submitted to ImageNet challenge, the GBD-v1 is modified. The modified GBD-Net structure has the following formulation:
\begin{align}
&\mathbf{h}_{i}^{1} = \sigma (\mathbf{h}^0_{i} \otimes \mathbf{w}_{i}^{1}+\mathbf{b}^{0,1}_{i}) + G_{i}^1  \bullet \sigma(\mathbf{h}_{i-1}^{1} \otimes \mathbf{w}_{i-1,i}^{1}+\mathbf{b}^1_{i}), \label{eqa:hi1_2} \\ 
&\mathbf{h}_{i}^{2} = \sigma(\mathbf{h}^0_{i} \otimes \mathbf{w}_{i}^{2}+\mathbf{b}^{0,2}_{i}) + G_{i}^2  \bullet \sigma(\mathbf{h}_{i+1}^{2} \otimes \mathbf{w}_{i,i+1}^{2}+\mathbf{b}^{2}_{i}), \label{eqa:hi2_2} \\
&\mathbf{h}_{i}^{3,m} = \max(\mathbf{h}_{i}^{1},\mathbf{h}_{i}^{2}),    \\
&\mathbf{h}_{i}^{3} =\mathbf{h}^0_{i} + \beta \mathbf{h}_{i}^{3,m},  \label{eqa:hi3_2} 
\end{align}
where $G_{i}^1$ and $G_{i}^2$ are defined in (\ref{eqa:gi1}) and (\ref{eqa:gi2}).
Fig. \ref{fig:net_ext} shows the modified GBD structure. The operations requried for obtaining $\mathbf{h}_{i}^{1}$ and $\mathbf{h}_{i}^{2}$ are the same as before. The main changes are in obtaining $\mathbf{h}_i^3$. The changes made are as follows.

First, in the previous GBD structure, $\mathbf{h}_{i}^{1}$ and $\mathbf{h}_{i}^{2}$ are concatenated and then convoled by filters  to produce the output $\mathbf{h}_{i}^{3}$. In the modified sturcture, a max pooling is used for merging the information from $\mathbf{h}_{i}^{1}$ and $\mathbf{h}_{i}^{2}$. This saves the memory and computation required by the convolution in the previous GBD structure.

Second, we add an identity mapping layer in the struture, which corresoponds to the $\mathbf{h}_{i}^{3} =\mathbf{h}^0_{i} + ...$ in (\ref{eqa:hi3_2}). The aim of the GBD structure is to refine the input $\mathbf{h}^0_{i}$ by using the messages from other contextual features. Since the parameters for the layers after the output $\mathbf{h}_{i}^{3}$ are from pretrained model, a drastic change of the output $\mathbf{h}_{i}^{3}$ from the input $\mathbf{h}_{i}^{0}$ would cause difficulty in training the layers after the layer $\mathbf{h}_{i}^{3}$ to adapt at the training stage. Therefore, this drastic change would lead to difficulty in learning a good model. When we train the previous GBD structure, careful initialization of the convolution parameters and the gate functions has to be done in order to learn well. For example, we have to set the gate function close 0 and the convolution parameter close to identity mapping for initialization. With the identity mapping layer, however, a simple initialization using the approach in \cite{Bengio2010Understanding} works well.

Third, a constant $\beta$ is multiplied with the merged messages $\mathbf{h}_{i}^{3,m}$ from other contextual regions. We empirically found that it improves detection accuracy by using $\beta$ to control the magnitude of the messages from other contextual features.

\subsection{Implementation details, training scheme, and loss function}
\label{sec:training}

For the state-of-the-art fast RCNN object detection framework, CNN is first pre-trained with the ImageNet image classification data, and then utilized as the initial point for fine-tuning the CNN to learn both object confidence scores $s$ and bounding-box regression offsets $t$ for each candidate box. 
Our proposed framework also follows this strategy and randomly initialize the filters in the gated bi-direction structure while the other layers are initialized from the pre-trained CNN. The final prediction on classification and bounding box regression is based on the representations $\mathbf{h}^{3}_i$ in equation (\ref{eqa:hi3}). For a training sample with class label $y$ and ground-truth bounding box offsets $\mathbf{v}=[v_1, v_2, v_3, v_4]$, the loss function of our framework is a summation of the cross-entropy loss for classification and the smoothed $L_{1}$ loss for bounding box regression as follows:

{\small
\begin{align}
L(y, t_y, \mathbf{v}, t_v ) =  L_{cls}(y, t_y)  + \lambda [y \geq 1] L_{loc}( \mathbf{v}, \mathbf{t}_v), \\
L_{cls}(y, t_y)  = -\sum_c \delta(y, c) \log t_{c}, \\
L_{loc}(\mathbf{v}, \mathbf{t}_v) =  \sum_{i=1}^{4}\text{smooth}_{L_{1}}(v_i- t_{v, i}),  \\
\text{smooth}_{L_{1}}(x) = 
\begin{cases}
0.5 x^{2} & \text{if } |x| \leq 1 \\
|x|-0.5 & otherwise
\end{cases},
\end{align}
}
\!\!\!where the predicted classification probability for class $c$ is denoted by $t_{c}$, and the predicted offset is denoted by $ \mathbf{t}_v=[t_{v, 1}, t_{v, 2}, t_{v, 3}, t_{v, 4}]$, $\delta(y, c)=1$ if $y=c$ and  $\delta(y, c)=0$ otherwise. $\lambda=1$ in our implementation.  Parameters in the networks are learned by back-propagation.

\subsection{Discussion}
\label{Sec:discussion}
Our GBD-Net builds upon the features of different resolutions and contexts. Its placement is independent of the place of roi-pooling. In an extreme setting, roi-pooling can be directly applied on raw pixels to obtain features of multiple resolutions and contexts, and in the meanwhile  GBD-Net can be placed in the last convolution layer for message passing. In this implementation, fast RCNN is reduced to RCNN where multiple regions surrounding a candidate box are cropped from raw pixels instead of feature maps.

\section{The object detection framework on ImageNet 2016}
\label{Sec:CGBD}
In this section, we describe object detection framework used for our submission to the 2016 ImageNet Detection Challenge. 

\subsection{Overview at the testing stage}
\begin{itemize}
  \item Step 1) Candidate box generation. An improved version of CRAFT in \cite{yang2016craft}  is used for generating the region proposal.
  \item Step 2) Box classification. The GBD-Net predicts the class of candidate boxes.
  \item Step 3) Average of multiple deep model outputs is used to improve the detection accuracy.
  \item Step 4) Postprocessing. The whole-image classification scores are  used as contextual scores for refining the detection scores. The bounding box voting scheme in \cite{gidaris2015object} is adopted for adjusting the box locations based on its neigbouring boxes. 
\end{itemize}

\subsection{Candidate box generation}  

We use two versions of object proposal. In early version of our method, we use the solutions published in \cite{yang2016craft} which is denoted as Craft-v2. In the final ImageNet submission, we further improve the results and denote it as Craft-v3. A brief review of Craft-v2 and details of Craft-v3 are described as follows.

\subsubsection{Craft-V1 and V2} In Craft \cite{yang2016craft}, the RPN \cite{ren2015faster} is extended to be a two-stage cascade structure, following the ``divide and conquer '' strategy in detection task. In the first stage, the standard RPN is used to generated about 300 proposals for each image, which is similar to the setting in \cite{ren2015faster}. While in the second stage, a two-category classifier is further used to distinguish objects from background. Specially in the paper, we use a two-category fast RCNN \cite{girshick2015fast}. It provides fewer and better localized object proposals than the standard RPN.  Craft-V1, which was used in our preliminary version \cite{zeng2016gated}, and Craft-V2 can be found it our early paper \cite{yang2016craft}.  Craft-V1 and Craft-V2 are only different in pre-training. Craft-V1 is pre-trained from 1000-class image classification, Craft-V2 is pre-trained from RPN \cite{ren2015faster}.

\subsubsection{Craft-v3} Compared with Craft-v2, the differences in Craft-v3 includes:
\begin{itemize}
   \item Random crop is used in model training, to ensure objects in different scales are roughly trained equally.
   \item Multi-scale pyramid is used in model testing, in order to improve recall of small objects.
   \item The positive and negative samples in RPN training are balanced to be 1:1. 
   \item LocNet \cite{gidaris2016locnet} object proposals are added, which we found are complementary to the Craft based proposals. 
\end{itemize}

Implementation details and experimental comparison can be found it Section \ref{Sec:proposalexp}.

\subsection{Box classification with GBD-Net}
The GBD-Net is used for predicting the object category of the given candidate boxes. The preliminary GBD-Net structure in \cite{zeng2016gated} was based on the BN-net. In the challenge we make the following modifications:
 \begin{itemize}
\item The baseline network is pretrained on ImageNet 1000-class data with object-centric labels without adapting to fas RCNN. In the challenge, we learn the baseline network with object-centric labels by adapting it to fas RCNN.
\item A ResNet with 269 layers is used as the baseline model for the best performing GBD-Net. 
\item The structure of GBD-Net is changed from GBD-v1 to GBD-v2, with details in Section \ref{Sec:GBD_ext}.
\end{itemize}

\subsection{Pretraining the baseline}
\subsubsection{The baseline ResNet-269 model}
The network structure of baseline ResNet with 269 layers is shown in Fig. \ref{fig:ResNet}. Compared with the ResNet \cite{he2016deep} with 152 layers, we simply increase the number of stacked blocks for conv3\_x, conv4\_x, and conv5\_x. The basic blocks adopt the identity mapping used in \cite{he2016identity}. At the training stage, the stochastic depth in \cite{huang2016deep}  is used. Stochastic depth is found to reduce training time and test error in \cite{huang2016deep}. For fast RCNN, we place the roi-pooling after the 234th layer, which is in the stacked blocks conv4\_x.

\subsubsection{Adapt the pretraining for roi-pooling}
\begin{figure*}[h]
\begin{center}
\includegraphics[width=0.97\linewidth]{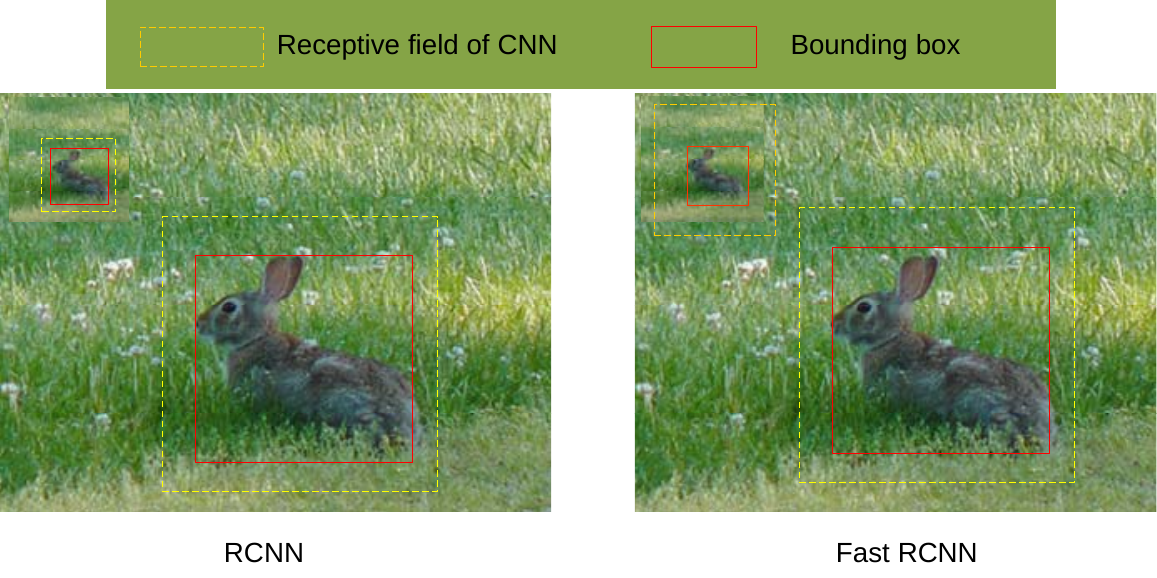}
\end{center}
\caption{The number of padded pixels changes as the box size changes for RCNN but is kept the same for fast RCNN.}
\label{fig:FieldSize}
\end{figure*}

\begin{figure*}[h]
\begin{center}
\includegraphics[width=0.97\linewidth]{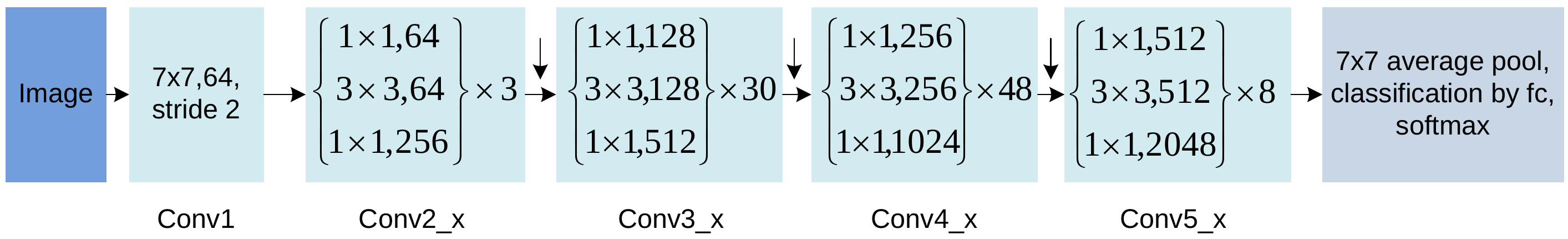}
\end{center}
\caption{Architecture for the baseline ResNet-269. Building blocks are the identity mapping blocks used in \cite{he2016identity}, with the numbers of blocks stacked. Downsampling is performed by conv3\_1, conv4\_1, and conv5\_1 with a stride of 2.}
\label{fig:ResNet}
\end{figure*}

Pretraining can be done on the ImageNet 1000-class image classification data by taking the whole image as the input of CNN, this is called image-centric pretraining. On the other hand, since bounding box labels are provided for these classes in ImageNet, the input of CNN can be obtained from warping image patches in the bounding boxes, which is called object-centric pretraining.
For the RCNN framework, it is found by our previous work \cite{ouyang2015deepid} that object-centric pretraining performs better than image-centric pretraining. For fast RCNN, however, we found that the CNN with object-centric pretraining does not perform better than the 1000-class image-centric pretraining. Take the BN-net as an exmaple, after finetuning on the ImageNet train+val1 data,  the BN-net from the image-centric pretraining has 49.4\% mAP on the ImageNet val2 while the BN-net from the object-centric pretraining has 48.4\% mAP.
This is caused by the following two reasons:
\begin{itemize}
\item Change from constant relative contextual region in RCNN to variant relative contextual region fast RCNN.
The receptive field of a neuron refers to the image region that influences the value of a neuron. In RCNN, the a contextual region with 0.2 the size of the bounding box is defined. Image patch within this contextual region is warped into a target size and then used as the input of CNN. Take BN-net for example, this target size is $224\times 224$. The CNN only takes  the visual cues in this contextual region into consideration. The number of pixels for the contextual region changes as the bounding box size changes. When pretraining the network for object-centric classification in \cite{ouyang2015deepid}, we adopted the RCNN scheme. The contextual region size changes as the bounding box changes.
For fast RCNN, however, the contextual region size is the same for bounding boxes of different sizes. In this case, the relative contextual region changes as bounding box size changes.
Take the BN-net with roi-pooling at inception (4d) for example. Because of the convolution and pooling in multiple layers, a neuron in inception (4d) has receptive field being 379.  Therefore, the contextual region with 378 pixels influences the neurons after roi-pooling for bounding box of any size. For a bounding box with size $224 \times 224$, the contextual region used in the roi-pooled features has 378 pixels. On the other hand, for a bounding box with size $112 \times 112$,  the contextual region used in the roi-pooled features also has 378 pixels. An example is shown in Fig. \ref{fig:FieldSize}. 
This is caused by the convolution on the whole image in fast RCNN compared with the convolution on image patch in RCNN. The   padded zeros in convolutional layers for the image patch in RCNN are replaced by contextual image region in fast RCNN. Therefore, the object-centric pretraining in \cite{ouyang2015deepid} for RCNN does not match fast RCNN in terms of relative contextual region.
\item Change from image warping in RCNN to feature warping in fast RCNN. In order to obtain features of the same size for candidate regions of different sizes, RCNN warp the candidate image region. Fast RCNN keeps the image size unchanged and uses roi-pooling for warping features.  Warping at image level for RCNN and roi-pooling at feature level are not equivalent operations.
\end{itemize}

In order to have the pretrained CNN aware of the differences mentioned above, we pretrain on object-centric task with roi-pooling layer included.  Let $(x_1, y_1, x_2, y_2)$ denote the bounding box of an object. When pretraining the BN-net with the roi-pooling layer, we include 32 pixels as the contextual region for this object. Therefore, the target box is $(x_{t,1}, y_{t,1}, x_{t,2}, y_{t,2}) = (x_1-32, y_1-32, x_2+32, y_2+32)$. To augment data, we randomly shake the target box as follows:
\begin{align}
\mathbf{b}_f &=(x_{f,1}, y_{f,1}, x_{f,2}, y_{f,2}) \label{eq:bf} \\
  &= (x_{t,1}+\alpha_1 W, y_{t,1}+\alpha_2 H, x_{t,2}+\alpha_3 W, y_{t,2}+\alpha_4 H), \nonumber
\end{align}
where $W=x_2-x_1+1$ and $H=y_2-y_1+1$ are respectively the width and height of the bounding box. $\alpha_1$, $\alpha_2$, $\alpha_3$, and $\alpha_4$ are randomly sampled from $[-0.1\ 0.1]$ independently. The image region within the box $\mathbf{b}_f$ in (\ref{eq:bf}) is warped into an image with shorter side randomly sampled from $\{300,350,400,450,500 \}$ and the longer side constrained to be no greater than 597. Batch size is set as 256 with other settings the same as BN-net. We observe 1.5\% mAP gain for BN-net and around 1.0\% mAP gain for ResNet-101 with this pretraining when compared with pretraining by image-centric classification. 
%For ResNet-269, however,  image-centric pretraining and object-centric pretraining are very close in performance. We tried more training configurations for BN-net than for ResNet-269. There is still possible performance gain in using object-centric pretraining for ResNet.

\subsection{Technical details on improving performance}
\label{Sec:improve}

\subsubsection{Multi-scale testing}
Multi-scale training/testing has been developed in \cite{girshick2015fast, he2014spatial} by selecting from a
feature pyramid. We only use the multi-scale input at the testing stage.
 With a trained model, we compute feature maps on an image pyramid, with the shorter side of the image being $\{400, 500, 600, 700, 800 \}$ and longer size being no greater than 1000. Roi-pooling and its subsequent layers are performed on
the feature map of one scale. We did not observe obvious improvement by averaging the scores of a bounding box using its features pooled from multiple scales.

\subsubsection{Left-right flip}
We adopt left-right flip at both the training stage and testing stage. At the training stage, the training images are augmented by flipping them. The candidate boxes are flipped accordingly. At the testing stage, an image and its corresponding candidate boxes are flipped and treated as the input of the CNN to obtained the detection scores for these boxes. The scores and estimated box locations from the original image and the flipped image are averaged.

\subsubsection{Bounding box voting} The bounding box voting scheme in \cite{gidaris2015object} is adopted. After finding the peaked box with the highest score on its neighborhood, the final object location is obtained by having each of the boxes that overlap with the peaked one by more than a threshold to vote for the bounding box location using its score as weight. This threshold is set as 0.5 for IoU.

\subsubsection{Non-maximum suppression (NMS) threshold}
For ImageNet, the NMS threshold was set as 0.3 by default. We empirically found 0.4 to be a better threshold. Setting this threshold from 0.3 to 0.4 provides 0.4-0.7\% mAP gain, with variation for different models.

\subsubsection{Global context}
From the pretrained image-centric CNN model, we finetune on the ImageNet detection data by treating it as an image classification problem. The 200-class image classification score is then used for combining with the 200-class object detection scores by weighted averaging. The weight are obtained by greedy search from the val1 data. 

\subsubsection{Model ensemble}
For model ensemble, 6 models are automatically selected by greedy search on ImageNet Det val2 from 10 models. The average of scores and bounding box regression results of these 6 models are used for obtaining the model averaging results.

\section{Experimental Results}
%This section reports the experimental results in Section \ref{Sec:GBD}, which is provided in the preliminary version \cite{zeng2016gated}.
\subsection{Implementation details}
\label{Sec:ImplementaionBN}
The  GBD-net is implemented based on the fast RCNN pipeline. The BN-net will be used for ablation study and our submission to the challenge is based on the ResNet with identity mapping \cite{he2016identity} and the model from the team of CU-DeepLink. 
%The exemplar implementation in Section \ref{sec:overview} and Fig. \ref{fig:examplar} is used in the experimental results if not specified. 
The gated bi-directional structure is added after the 6th inception module (4d) of BN-net and after the 234th layer for ResNet-269. In the GBD-Net, layers belonging to the baseline networks are initialized by these baseline networks pre-trained on the ImageNet 1000-class classification and localization dataset. The parameters in the GBD layers as shown in Fig.  \ref{fig:net}, which are not present in the pre-trained models, are randomly initialized when finetuning on the detection task. In our implementation of GBD-Net, the feature maps $\mathbf{h}_{i}^{n}$ for $n=1, 2, 3$ in (\ref{eqa:hi1})-(\ref{eqa:hi3}) have the same width, height and number of channels as the input $\mathbf{h}_{i}^{0}$  for $i=1, 2,3,4$.

We evaluate our method on three public datasets, ImageNet object detection dataset \cite{ILSVRCarxiv14}, Pascal VOC 2007 dataset \cite{Everingham:PacalVOC} and Microsoft COCO object detection dataset \cite{lin2014microsoft}. Since the ImageNet object detection task contains a sufficiently large number of images and object categories to reach a conclusion, evaluations on component analysis of our training method are conducted on this dataset. This dataset has 200 object categories and consists of three subsets. i.e., train, validation and test data. In order to have a fair comparison with other methods, we follow the same setting in \cite{girshick2014rich} and split the whole validation subset into two subsets, val1 and val2.  The network finetuning step uses training samples from train and val1 subsets. The val2 subset is used for evaluating components and the performance on test data is from the results submitted to the ImageNet challenge. Because the input for fast RCNN is an image from which both positive and negative samples are sampled, we discard images with no ground-truth boxes in the val1. Considering that lots of images in the train subset do not annotate all object instances, we reduce the number of images from this subset in the batch. For all networks, the learning rate and weight decay are fixed to 0.0005 during training, the batch size is 192. We use batch-based stochastic gradient descent to learn the network and the batch size is 192. The overhead time at inference due to gated connections is less than 40\%.

\subsection{Overall performance}

\subsubsection{PASCAL VOC2007 dataset}
It contains 20 object categories. Following the most commonly used approach in \cite{girshick2014rich}, we finetune the BN-net with the 07+12 trainval set and evaluate the performance on the test set. Our GBD-net obtains 77.2\% mAP while the baseline BN+FRCN is only 73.1\%.

\subsubsection{Microsoft COCO object detection dataset}
We use MCG \cite{arbelaez2014multiscale} for region proposal and report both the overall AP and AP$^{50}$ on the closed-test data. The baseline BN+FRCN implemented by us obtains 24.4\% AP and 39.3\% AP$^{50}$, which is comparable with Faster RCNN (24.2\%  AP) on COCO detection leadboard. With our proposal gated bi-directional structure, the network is improved by 2.6\%  AP points and reaches 27.0\%  AP and 45.8\%  AP$^{50}$, which further proves the effectiveness of our model.

\subsubsection{ImageNet object detection dataset}
We compare our framework with several other state-of-art approaches \cite{girshick2014rich,szegedy2015going,ioffe2015batch,ouyang2015deepid,yan2015object,he2016deep}. The mean average precision for these approaches are shown in Table \ref{table:overall}.  Our work is trained using the provided data of ImageNet. Compared with the published results and recent results in the provided data track on ImageNet 2015 challenge, our single model result performs better than the ResNet \cite{he2016deep} by 4.5\% in mAP for single-model result. 

Table \ref{Tab:ILSVRC16} shows the experimental results for UvA, GoogleNet, ResNet, which are best performing approaches in the ImageNet challenge 2013, 2014 and 2015 respectively. The top-10 approaches attending the challenge 2016 are also shown in Table \ref{Tab:ILSVRC16}. Our approach has the similar mAP as Hikvision in single model and performs better for averaged model. Among the 200 categories, our submission wins 109 categories in detection accuracy.

%The BN-net on Fast RCNN implemented by us is our baseline, which is denoted by BN+FRCN. From the table, it can be seen that BN-net with our GBD-Net has 5.1\% absolute mAP improvement compared with BN-net. We also report the performance of feature combination method as opposed to gated connections, which is denoted by BN+FC+FRCN. It uses the same four region features as GBD-net by simple concatenation and  obtains 47.3\% mAP, while ours is 51.4\%.
 
\begin{table*}
\centering
{\small
\caption{Object detection mAP (\%) on ImageNet val2 for state-of-the-art approaches with single model (sgl) and averaged model (avg).}
\begin{tabular}{cccccccc}
\hline
appraoch& RCNN & Berkeley & GoogleNet &  DeepID-&Superpixel &ResNet&Ours \\
&\cite{girshick2014rich}&\cite{girshick2014rich}&\cite{szegedy2015going}&Net\cite{ouyang2015deepid}&\cite{yan2015object}&\cite{he2016deep}&\\
\hline
val2(sgl)&31.0&33.4&38. 5&48.2&42.8&60.5&65\\
val2(avg)&n/a&n/a&40.9&50.7&45.4&63.6&68\\
\hline
\end{tabular}
}
\label{table:overall}
\end{table*}

\begin{table*}
\setlength{\tabcolsep}{1.5pt}
\centering
\caption{Object detection mAP (\%) on ImageNet for the approaches attending the ImageNet challenge with single model (sgl) and averaged model (avg) when tested on the val2 data and Test data without using external data for training.}
\label{Tab:ILSVRC16}
\begin{tabular}{ccccccccccccccc}
\hline
Year       & 2013         & 2014      & 2015   & 2016 & 2016    & 2016    & 2016       & 2016 & 2016    & 2016           & 2016  & 2016      & 2016 \\
Team       & UvA & GoogleNet & ResNet & VB   & Faceall & MIL\_UT & KAIST-SLSP & CIL  & 360+MCG & Trimps & NUIST & Hikvision & Ours \\
\hline
val2 (sgl) & -            & 38.8      & 60.5   & -    & 49.3    & -       & -          & -    & -       & -              & -     & 65.1      & 65   \\
val2 (avg) & -            & 44.5      & 63.6   & -    & 52.3    & -       & -          & -    & -       & -              & -     & 67        & 68   \\
Test (avg) & -            & 38        & 62.1   & 48.1 & 48.9    & 53.2    & 53.5       & 55.4 & 61.6    & 61.8           & 60.9  & 65.2      & 66.3 \\
Test (sgl) & 22.6         & 43.9      & 58.8   & -    & 46.1    & -       & -          & -    & 59.1    & 58.1           & -     & 63.4      & 63.4
 \\
\hline
\end{tabular}
\end{table*}

\subsection{Investigation on different settings in GBD-v1}
\label{Sec:GBDexpv1}
\subsubsection{Investigation on placing roi-pooling at different layers}
The placement of roi-pooling is independent of the placement of the GBD-Net. Experimental results on placing the roi-pooling after the image pixels and after the 6th inception module are reported in this section.
If the roi-pooling is placed after the 6th inception module (4d) for generating features of multiple resolutions, the model is faster in both training and testing stages. If the roi-pooling is placed after the image pixels for generating features of multiple resolutions, the model is slower because the computation in CNN layers up to the 6th inception module cannot be shared.
Compared with the GBD-Net placing roi-pooling after the 6th inception module with mAP 48.9\%, the GBD-Net placing the roi-pooling after the pixel values with mAP 51.4\% has better detection accuracy. This is because the features for GBD-Net are more diverse and more complementary to each other when roi-pooling is placed after pixel values.

%\begin{table}
%\centering
%{\small
%\caption{Object detection mAP (\%) on ImageNet val2 for BN-net and BN-net with our GBD-Net when roi-pooling is placed on inception (4d) or pixels.}
%\begin{tabular}{c|c|cc}
%\hline
%\multirow{2}{*}{BN-net} 
%&\multicolumn{2}{|c}{BN-net with GBD-net}& \\
%\cline{2-4}
%& roi-pooling on inception (4d) & roi-pooling on pixels \\
%\hline
%46.3 & 48.9 & 51.4 \\
%\hline
%\end{tabular}
%}
%\label{table:roiRes}
%\end{table}  

\subsubsection{Investigation on gate functions}
Gate functions are introduced to control message passing for individual candidate boxes. Without gate functions, it is hard to train the network with message passing layers in our implementation. It is because nonlinearity increases significantly by message passing layers and gradients explode or vanish, just like it is hard to train RNN without LSTM (gating). In order to verify it, we tried different initializations. The network with message passing layers but without gate functions has 42.3\% mAP if those message passing layers are randomly initialized. However, if those layers are initialized from a well-trained GBD-net, the network without gate functions reaches 48.2\% mAP. Both two results also show the effectiveness of gate functions.

\subsubsection{Investigation on using different feature region sizes}
\begin{table*}
\centering
\caption{Detection mAP (\%) for features with different padding values $p$ for GBD-Net-v1 using BN-net as the baseline. Craft-V1 is used for region proposal. Different $p$s leads to different resolutions and contexts.}
\begin{tabular}{c|c|c|c|c|c|c|c|c}
\hline
\multirow{2}{*}{Padding value $p$} & \multicolumn{4}{c}{Single resolution}& \multicolumn{4}{|c}{Multiple resolutions} \\
\cline{2-9}
& -0.2 & 0.2 & 0.8 & 1.7 & -0.2,0.2 & 0.2+1.7 & -0.2+0.2+1.7 & -0.2+0.2+0.8+1.7 \\
\hline
mAP&46.3&46.3&46.0&45.2&47.4&47.0&48.0&48.9\\
\hline
\end{tabular}
\label{table:scale}
\end{table*} 

The goal of our proposed gated bi-directional structure is to pass messages among features with different resolutions and contexts. In order to investigate the influence from different settings of resolutions and contexts, we conduct a series of experiments. In these experiments, features of a particular padding value $p$ is added one by one.The experimental results for these settings are shown in Table \ref{table:scale}. When single padding value is used, it can be seen that simply enlarging the support region of CNN by increasing the padding value $p$ from 0.2 to 1.7 does harm to detection performance because it loses resolution and is influenced by background clutter. On the other hand, integrating features with multiple resolutions and contexts using our GBD-Net substantially improves the detection performance as the number of resolutions/contexts increases. Therefore, with the GBD-Net, features with different resolutions and contexts help to validate the existence of each other in learning features and improve detection accuracy. 

\subsubsection{Investigation on combination with multi-region}
This section investigates experimental results when combing our gated bi-directional structure with the multi-region approach. We adopt the simple straightforward method and average the detection scores of the two approaches. The baseline BN model has mAP 46.3\%. With our GBD-Net the mAP is 48.9\%. The multi-region approach based on BN-net has mAP 47.3\%. The performance of combining our GBD-Net with mutli-region BN is 51.2\%, which has 2.3\% mAP improvement compared with the GBD-Net and 3.9\% mAP improvement compared with the multi-region BN-net. This experiment shows that the improvement brought by our GBD-Net is complementary to the multi-region approach in \cite{gidaris2015object}. 

\subsection{Investigation on GBD and GBD-v2}
This section investigates the experimental results for the GBD in Section \ref{sec:net} and the GBD-v2 introduced in Section \ref{Sec:GBD_ext}. For the same BN-net, the GBD-v1 introduced in Section \ref{Sec:GBD} improves the BN-net by 2.6\% in mAP. The GBD-v2 structure introduced in Section \ref{Sec:GBD_ext} improves mAP by 3.7\%. Therefore, the GBD-v2 is better in improving the detection accuracy. Since GBD-v2 has better performance and is easier to train, we use the GBD-v2 as the default GBD structure in the following part of this paper if not specified. Since the components investigated in Section \ref{Sec:GBDexpv1} are not different for GBD-v1  and GBD-v2, we directly adopt the settings found to be effective in GBD-v1 for GBD-v2. 
In GBD-v2, roi-pooling is placed at the module (4d) for BN-net and the 234th layer for ResNet-269. Gate function is used. We use feature regions with padding values $p=$ -0.2, 0.2, 0.8, 1.7.

\subsection{Investigation on other components in the detection framework}
In the component-wise study, none of the techniques in Section \ref{Sec:improve} is included if not specified. We adopt the left-right flip at the the training stage for data augmentation for all of the evaluated approaches but did not use flip at the testing stage if not specified.

%
% \section{Experimental Results for the ImageNet Challenge Submission}
% \label{Sec:ExpImageNet16}
%This section reports the experimental results related to the crafted detection framework introduced in Section \ref{Sec:CGBD}. 

%Our detection framework is implemented based on the fast RCNN pipeline. The implementation and experimental setup in Section \ref{Sec:ImplementaionBN} are used in the experimental results if not specified. 
% The val2 dataset will be used for evaluating components and the performance on test data is from the results submitted to the ImageNet challenge. 
% 

%\subsection{Overal results}

%\subsection{Component-wise study}

\subsubsection{Region proposal} \label{Sec:proposalexp}
We list the improvements on top of our early work Craft-v2 \cite{yang2016craft} in Table \ref{Table:GBD_proposal}. Random crop in training and multi-scale testing also helps and they leads to 0.74\% and 1.1\% gain in recall, respectively.  In multi-scale training, we want to keep the distribution of image size after log operation is uniform. To this end, in each iteration, for each image, we randomly select a scale number $r$ in range of $[16,512]$ and randomly select an object in this image with the length $l$. Then the resize scale is set to be $l/r$. 
This multi-scale training improves recall by 0.7\%.

In the testing stage, we densely resize the longer side of each image to $2800\times 2^{(-9:0)}$ and found that it is necessary to achieve high enough recall for objects range in $[20,50]$ pixels for longer side. 

To balance the positive and negative samples, we implement a new multi-GPU implementation, where 50\% of the GPUs only train positive samples while the other 50\% GPUs only train the negative ones. Balancing the positive and negative samples to $1:1$ in training leads to 0.6\% gain in recall.

We use 150 proposals for each image generated by the Craft framework. We combine two methods to get 300 proposals for each image and named it as Craft-v3. For the Craft-v3, the recall on ImageNet val2 is 95.30\%, and the average recall \footnote{please refer more details of the proposal average recall to \cite{hosang2016makes}.} is 1.59. 

Use the same BN-net as the baseline on ImageNet val2, Craft-V1, V2 and V3 have mAP 46.3,  48.4, and 49.4 respectively.
Compared to the Craft-v2 on ImageNet val2, the Craft-v3 leads to 1.9\% gain in final detection AP for ResNet-269+GBD-Net.

\begin{table*}[]
\setlength{\tabcolsep}{2.5pt}
\centering
\caption{Recall rate on ImageNet val2 with different components in proposal generation.}
\label{Table:GBD_proposal}
\begin{tabular}{cccccc}
\hline
Components. &baseline (Craft-v2) \cite{yang2016craft} &  +Random Crop training & +Multi-scale testing & +Balancing positive and negative samples &+ensemble\\
\hline
Recall @ 150 proposals   & 92.37\%            &  93.11\%         & 94.21\%        & 94.81\%          &              \\
Recall @ 300 proposals   &                    &                  &                &                  & 95.30\%        \\
\hline 
\end{tabular}
\end{table*}

\subsubsection{Pretraining scheme}
There are three pretraining schemes evaluated in the experimental results shown in Table \ref{Table:prtrain}. The BN-net is used as the network for evaluation. The pretraining on the 1000-class image-centric classification task without using the box annotation of these objects, which is denoted by \emph{image} in Table  \ref{Table:prtrain}. The pretraining on the 1000-class object-centric classification task using the box annotations without including the roi-pooling at the pretraining stage, which is denoted by  \emph{object w/o roi}  in Table  \ref{Table:prtrain}. The pretraining for the 1000-class object-centric classification task with the box annotations and with the roi-pooling included at the pretraining stage, which is denoted by \emph{object+roi} in Table  \ref{Table:prtrain}. Without using the roi-pooling at the pretraining stage, the object-centric pretraining performs worse than image-centric pretraining. The inclusion of roi-pooling at the pretraining stage improves the effectiveness of object centric pretraining for finetuning in object detection, with 1.5\% increase in absolute mAP.

\begin{table*}
\centering
\caption{Object detection mAP (\%) on ImageNet val2 for BN-net using different pretraining schemes.}
\label{Table:prtrain}
\begin{tabular}{cccc}
\hline
Pretraining scheme & Image    & Object w/o roi & Object+roi  \\
Region proposal    & Craft-V2 & Craft-V2  & Craft-V2 \\
\hline
mAP                & 49.1 & 48.4      & 49.9   \\
\hline
\end{tabular}
\end{table*}

\subsubsection{The GBD structure}
We evaluate the GBD structure for different baseline models. Fig. \ref{Table:GBD_new1} shows the experimental results for different baseline networks with the GBD structure. The GBD structure introduced in Section \ref{Sec:GBD_ext} improves the BN-net by 3.7\% in mAP for Craft-V2 and by 4.2\% in mAP for Craft-V3. With GBD and the better region proposal, BN-net+GBD with mAP 53.6\% is close to ResNet-152 with mAP 54\% in detection accuracy. The modified GBD structure improves the mAP by 2.5\% and 2.2\% for ResNet-152 and ResNet-269 respectively.

\begin{table*}
\centering
\caption{Object detection mAP (\%) on ImageNet val2 for different baseline networks with the GBD-v2.The \emph{+ new GBD} denotes the use of the modified GBD structure in Fig. \ref{fig:net_ext} and introduced in Section \ref{Sec:GBD_ext}. }
\label{Table:GBD_new1}
\setlength{\tabcolsep}{1.5pt}
\begin{tabular}{ccccc}
\hline 
Baseline network & BN-net        & BN-net            & ResNet-152 & ResNet-269 \\
Region proposal  & Craft-V2       & Craft-V3        & Craft-V2   & Craft-V2   \\
Pretrain             & Object w/o roi & Object w/o roi   & Image      & Image    \\
\hline Without GBD      & 48.4      & 49.4                  & 54         & 56.6       \\
+ new GBD        & 52.1    & 53.6                  & 56.5       & 58.8       \\
\hline 
%Baseline network & BN-net          & BN-net          & ResNet-152 & ResNet-269 \\
%Region proposal  & Craft-V2        & Craft-V3        & Craft-V2   & Craft-V2   \\
%Pretrain             & object w/o roi & object w/o roi & image      & image    \\
%\hline Without GBD      & 48.1            & 50.1              & 54         & 56.6       \\
%+ new GBD        & 52.1            & 53.6            & 56.5       & 58.8       \\
\end{tabular}
\end{table*}

It is mentioned in Section \ref{Sec:GBD_ext} the magnitude of the messages from other contextual features influences the detection accuracy. Table \ref{Table:GBD_new2} shows the experimental results for messages with different magnitudes. In the experiments, the BN-net pretrained with bounding box label without roi-pooling is used as the baseline model. It can be seen that the scalar $\beta$ has the best performance when it is 0.1. Setting $\beta$ to be 1, i.e. not scaling messages, results in 1.6\% mAP drop.

\begin{table*}[]
\setlength{\tabcolsep}{2.5pt}
\centering
\caption{Object detection mAP (\%) on ImageNet val2 for BN-net using GBD structure with different scale factor $\beta$ in controlling the magnitude of message.}
\label{Table:GBD_new2}
\begin{tabular}{cccccc}
\hline 
Network config. &$\beta=1$ &  $\beta=0.5$ &  $\beta=0.2$ &  $\beta=0.1$ & $\beta$=0 ( without GBD) \\
Pretrain        & Object w/o roi     & Object w/o roi       & Object w/o roi       & Object w/o roi       & Object w/o roi          \\
Region proposal        & Craft-V3            & Craft-V3              & Craft-V3              & Craft-V3              & Craft-V3                 \\
mAP on val2     &  52                   & 53.2                  & 53.3                  & 53.6                  & 49.4        \\
\hline 
\end{tabular}
\end{table*}

\subsubsection{Baseline deep models}
In this section, we evaluate the influence of baseline deep models for the detection accuray on the ImageNet.
Table \ref{Table:Baseline} shows the experimental results for different baseline network structures. All models evaluated are pretrained from ImageNet 1000-class training data without using the bounding box label.  None of the model uses the stochastic depth \cite{huang2016deep} at the finetuning stage. If stochastic depth is included, it is only used at the pretraining stage.
From the results in \ref{Table:Baseline}, it can be seen that ResNet-101 with identity mapping \cite{he2016identity} and stochastic depth has 1.1\% mAP improvement compared with the ResNet-101 without them. Because of time limit and the evidence in ResNet-101, we have used the stochastic depth and identity mapping for the ResNet-269 baseline model.

\begin{table*}[]
\setlength{\tabcolsep}{2pt}
\centering
\caption{Object detection mAP (\%) on ImageNet val2 for different baseline deep models. All models are pretrained from ImageNet 1000-class classification data without using the bounding box label. None of the approaches introduced in Section \ref{Sec:improve} are used. `+I' denotes the use of identity mapping  \cite{he2016identity}. `+S' denotes the use of stochastic depth \cite{huang2016deep}. }
\label{Table:Baseline}
\begin{tabular}{cccccccl}
\hline
Net structure           & BN-net                  & ResNet-101          & ResNet-101+I+S                       & ResNet-152          & ResNet-269+I+S & Inception-V5                   & CU-DeepLink \\
                   & \cite{ioffe2015batch} & \cite{he2016deep} & \cite{he2016identity, huang2016deep} & \cite{he2016deep} &            & \cite{szegedy2015rethinking} &          \\
\hline
Pretraining scheme & Image                   & Image               & Image                                  & Image               & Image      & Image                          & Image    \\
Region proposal           & Craft-V2                & Craft-V2            & Craft-V2                               & Craft-V2            & Craft-V2   & Craft-V2                       & Craft-V3 \\
Mean AP                & 49.9                    & 52.7                & 53.8                                   & 54                  & 56.6       & 53.3                           & 57.2     \\
\hline
\end{tabular}
\end{table*}

\subsubsection{Model ensemble}
For model ensemble, we have used six models and the averge of their scores are used as the result for model ensemble. As shown in Table \ref{Table:modelavg}, these models vary in baseline model, pretraining scheme, use of GBD or not and region proposal for training the model. Note that the region proposal for training could be different, but they are tested using the same region proposal. Without context, the averaged model has mAP 66.9\%. With global contextual scores, the model has mAP 68\%.

\begin{table*}[]
\centering
\caption{Models used in the model ensemble, global contextual scores are not used in the results for these models.}
\label{Table:modelavg}
\begin{tabular}{ccccccc}
\hline
Model denotation        & 1            & 2          & 3            & 4          & 5         & 6           \\
Baseline model          & ResNet-269   & ResNet-269 & ResNet-269   & ResNet-269 & CU-DeepLink     & ResNet-101  \\
Use GBD               & \checkmark   & \checkmark & \checkmark   & \checkmark &           &             \\
Pretraining scheme      & Object + roi & Image      & Object + roi & Image      & Image     & Image       \\
Region proposal for training & Craft-V3     & Craft-V2   & Craft-V3     & Craft-V2   & Craft-V3  & Craft-V3    \\
%Single model mAP        & 63.5         & 62.55      & 63.44        & 62.05      & 62.33     & 62.38       \\
Averaged model         & 1            & 1+2        & 1+2+3        & 1+2+3+4    & 1+2+3+4+5 & 1+2+3+4+5+6 \\
 \hline
Average mAP            & 63.5         & 64.8       & 65.5         & 66         & 66.8     & 66.9     \\
\hline
\end{tabular}
\end{table*}

\subsubsection{Components improving performance}
\label{Sec:exp_techImprove}
Table \ref{Table:Results} summarizes experimental results for the components that improve the performance. The baseline ResNet-269 has mAP 56.6\%. With GBD-net, the mAP is 58.8\%. Changing the region proposal from Craft-v2 to Craft-v3 improves the mAP to 60.7\%. In the experimental results for the settings above, single-scale testing is used, in which the shorter side of the is constrained to be no greater than 600 and the longer is constrain to be  no greater than 700 at the testing and training stage. When the same model is used for multi-scale testing, we set scales to be [400, 500, 600, 700, 800] and the longer side constrain to be no greater than 1000. This multi-scale testing  provides 1.3\% mAP improvement. Left-right flip provides 0.7 mAP gain. Bounding box voting leads to 1.3 \% mAP gain. Changing the NMS threshold from  0.3 to 0.4 leads to 0.4 mAP gain. The use of context provides 1.3\% mAP improvement. The final single model result has 65\% mAP on the val2 data. Ensemble of six models improves the mAP by 3\% and the final result has 68\% mAP.

\begin{table*}[]
\centering
\caption{Summary of the components that lead to the final submission.}
\label{Table:Results}

\begin{tabular}{c|ccccccccc|c}
\hline
ResNet-269          & \checkmark & \checkmark & \checkmark & \checkmark & \checkmark & \checkmark & \checkmark & \checkmark & \checkmark & ResNet-269          \\
Craft-v2            & \checkmark & \checkmark &            &            &            &            &            &            &            &                     \\
Craft-v3            &            &            & \checkmark & \checkmark & \checkmark & \checkmark & \checkmark & \checkmark & \checkmark & Craft-v3            \\
Use GBD             &            & \checkmark & \checkmark & \checkmark & \checkmark & \checkmark & \checkmark & \checkmark & \checkmark & Use GBD             \\
Multi-scale testing &            &            &            & \checkmark & \checkmark & \checkmark & \checkmark & \checkmark & \checkmark & Multi-scale testing \\
Left-right flip     &            &            &            &            & \checkmark & \checkmark & \checkmark & \checkmark & \checkmark & Left-right flip     \\
Box voting          &            &            &            &            &            & \checkmark & \checkmark & \checkmark & \checkmark & Box voting          \\
NMS threshold       &    0.3         &    0.3         & 0.3        & 0.3        & 0.3        & 0.3        & 0.4        & 0.4        & 0.4        & NMS threshold       \\
Context             &            &            &            &            &            &            &            & \checkmark & \checkmark & Model ensemble      \\
Model ensemble      &            &            &            &            &            &            &            &            & \checkmark & Context             \\
mAP gain            & -          & 2.2        & 1.9        & 1.3        & 0.7        & 1.3        & 0.4        & 1.3        & 3          & mAP gain            \\
mAP                 & 56.6       & 58.8       & 60.7       & 62         & 62.7       & 63.3       & 63.7       & 65         & 68         & mAP                
   \\
\hline
\end{tabular}

\end{table*}

\subsection{Analysis of false positive types}
\begin{figure*}
\begin{center}
   \includegraphics[width=0.4\linewidth]{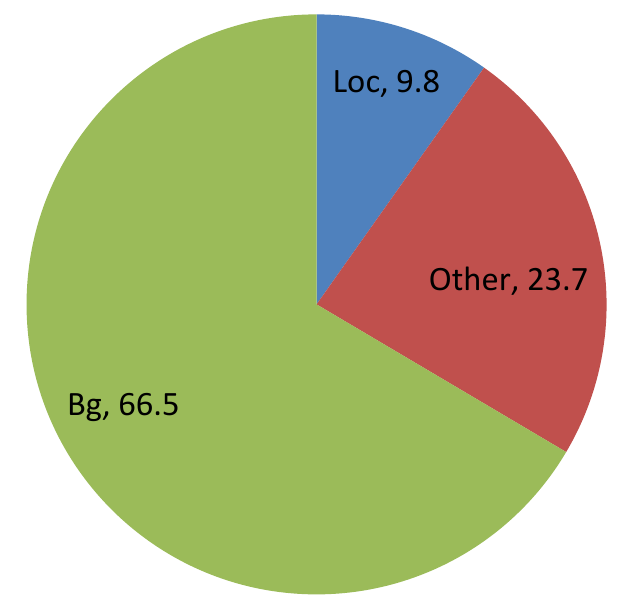}
\end{center}
   \caption{Fraction of high-scored false positives on ImageNet Val2 that are due to poor localization (Loc), confusion with other objects (Other), or confusion with background or unlabeled objects (Bg)}
\label{fig:fp}
\end{figure*}

Fig. \ref{fig:fp} shows the fraction of false positives on ImageNet Val2 that are caused by confusion with background, poor localization and confusion with other objects. It can be seen that, the majority of false positives are from background, which is different from the results in \cite{ouyang2015deepid} for Pascal VOC, where the majority of false positives are from poor localization. This is possibly from a better region proposal used in our approach.

%\subsection{Results analysis}
%In this section, we use the best performing result with 68.0\% on val2 for analyzing the properties of this detector. 

\section{Conclusion}
In this paper, we propose a gated bi-directional CNN (GBD-Net) for object detection. In this CNN, features of different resolutions and support regions pass messages to each other to validate their existence through the bi-directional structure. And the gate function is used for controlling the message passing rate among these features. Our GBD-Net is a general layer design which can be used for any network architecture and placed after any convolutional layer for utilizing the relationship among features of different resolutions and support regions.
The effectiveness of the proposed approach is validated on three object detection datasets, ImageNet, Pascal VOC2007 and Microsoft COCO. 

\section{Acknowledgment}
This work is supported by SenseTime Group Limited and the General Research Fund sponsored by the Research Grants Council of Hong Kong (Project Nos. CUHK14206114, CUHK14205615, CUHK417011, and CUHK14207814).

\clearpage

\bibliographystyle{splncs}
\bibliography{./PME}
%
%% if you will not have a photo
%\begin{biographynophoto}{John Doe}
%\end{biographynophoto}
%
%% insert where needed to balance the two columns on the last page
%%\newpage
%
%\begin{biographynophoto}{Jane Doe}
%\end{biographynophoto}


%\vfill


\end{document}